# Anomaly Detection in Double-entry Bookkeeping Data by Federated Learning System with Non-model Sharing Approach

Sota Mashiko[a], Yuji Kawamata[b], Tomoru Nakayama[c], Tetsuya Sakurai[d], Yukihiko Okada[e]

[a] Graduate School of Science and Technology, University of Tsukuba, Japan.
mashiko.sota.qd@alumni.tsukuba.ac.jp, 0009-0005-3638-6624

[b] Center for Artificial Intelligence Research, Tsukuba Institute for Advanced Research, University of Tsukuba, Japan. yjkawamata@gmail.com, 0000-0003-3951-639X

[c] Graduate School of Science and Technology, University of Tsukuba, Japan.
nakayama.tomoru.tkb_en@u.tsukuba.ac.jp

[d] Institute of Systems and Information Engineering, University of Tsukuba, Tsukuba, Japan. / Center for Artificial Intelligence Research, Tsukuba Institute for Advanced Research, University of Tsukuba, Tsukuba, Japan. sakurai@cs.tsukuba.ac.jp, 0000-0002-5789-7547

[e] Institute of Systems and Information Engineering, University of Tsukuba, Tsukuba, Japan. / Center for Artificial Intelligence Research, Tsukuba Institute for Advanced Research, University of Tsukuba, Tsukuba, Japan. okayu@sk.tsukuba.ac.jp, 0000-0003-4903-4191

**Corresponding Author**: Yuji Kawamata, yjkawamata@gmail.com, 1-1-1 Tennodai, Tsukuba, Ibaraki 305-8573, Japan

## Highlights

- An anomaly detection audit framework using data collaboration analysis is provided.
- The method requires no connection of devices storing raw data to external networks.
- Only a single round of communication is needed for training.
- It was validated on real journal entry data from multiple organizations.
- High performance under realistic non-i.i.d. auditing conditions was obtained.

## Abstract

Anomaly detection is crucial in financial auditing, and effective detection requires large volumes of data from multiple organizations. However, journal entry data is highly sensitive, making it infeasible to share them directly across audit firms. To address this challenge, journal entry anomaly detection methods based on model share-type federated learning (FL) have been proposed. These methods require multiple rounds of communication with external servers to exchange model parameters, which necessitates connecting devices storing confidential data to external networks—a practice not recommended for sensitive data such as journal entries. To overcome these limitations, a novel anomaly detection framework based on data collaboration (DC) analysis, a non-model share-type FL approach, is proposed. The method first transforms raw journal entry data into secure intermediate representations via dimensionality reduction and then constructs a collaboration representation used to train an anomaly detection autoencoder. Notably, the approach does not require raw data to be exposed or devices to be connected to external networks, and the entire process needs only a single round of communication. The proposed method was evaluated on both synthetic and real-world journal entry data collected from eight healthcare organizations. The experimental results demonstrated that the framework not only outperforms the baseline trained on individual data but also achieves higher detection performance than model-sharing FL methods such as FedAvg and FedProx, particularly under non-i.i.d. settings that simulate practical audit environments. This study addresses the critical need to integrate organizational knowledge while preserving data confidentiality, contributing to the development of practical intelligent auditing systems.

1. Introduction

In recent years, research on anomaly detection in the financial and accounting domains has progressed rapidly. The anomalies (frauds) under investigation are broadly categorized into external fraud (e.g., credit card fraud, insurance claim fraud, and loan fraud) and internal fraud (e.g., financial statement fraud and money laundering) (Hernandez et al., 2024). Among internal fraud studies, those focusing on financial statement manipulation have been identified as a particularly important research theme (Hernandez et al., 2024), and several studies have been published (e.g., Aftabi et al., 2023; Cai & Xie, 2024).

As part of audit standards, journal entry data play a pivotal role in financial statement audits (Debreceny & Gray, 2010). Journal entry data, which are recorded according to the rules of double-entry bookkeeping, comprise the voluminous daily transactions of an enterprise, making it impractical for auditors to inspect every entry manually. Consequently, computer-assisted audit techniques (CAAT) are often employed to extract and analyze these data digitally, screening suspicious transactions via a procedure known as "Journal Entry Testing." However, as these techniques typically rely on static rules, they often exhibit high false-positive rates (Schultz & Tropmann, 2020). In recent years, numerous anomaly detection methods based on machine learning and deep learning (DL) have been proposed (e.g., Bay et al., 2006; Schreyer et al., 2017; Zupan et al., 2020; Wei et al., 2024).

Such models require ample data volume to achieve high accuracy. Additionally, auditing firms accumulate industry-specific expertise by auditing multiple clients within the same sector, thereby improving both audit efficiency and quality (Hogan & Debra, 1999). These considerations suggest that integrating journal entry data obtained from several companies within the same industry could enable the development of more sophisticated anomaly detection methods. However, accounting data are highly confidential, making companies and auditing firms unwilling to share them directly. Consequently, approaches that preserve client data confidentiality while simultaneously consolidating knowledge across multiple organizations should be developed (Kogan & Yin, 2021).

To perform anomaly detection on such distributed confidential data while preserving privacy, federated learning (FL) (McMahan et al., 2017) has been increasingly adopted. For example, in credit card fraud-detection, models that combine optimization algorithms with FL (Reddy et al., 2024) and models that integrate graph neural networks (GNN) with FL (Tang & Liang, 2024) have been proposed. In the auditing domain, Schreyer et al. (2022) developed an FL-based framework for anomaly detection on journal entry data, presenting a method to construct industry-specific detection models across multiple organizations without sharing confidential data.

However, these methods assume that confidential data remain continuously connected to the internet. By contrast, highly sensitive datasets such as financial records are recommended to be managed in "air gap" environments—isolated from any external network (Guri, 2024)—and for data demanding the utmost confidentiality, such as journal entry data, approaches predicated on always-on connectivity are not advisable in practical settings.

In this study, a novel FL-based anomaly detection method for journal entry data—grounded in data collaboration (DC) analysis (Imakura & Sakurai, 2020)—that leverages distributed confidential data without requiring direct internet connectivity to raw data, is proposed (Figure 1). Unlike model share-type FL approaches, our method adopts a non-model share-type FL framework: each organization reduces the dimensionality of its raw data to produce intermediate representations, and only these representations are shared. This design ensures privacy while enabling collaborative model training. Moreover, the entire learning process needs only a single round of communication, promising a significant reduction in communication overhead compared with model share-type FL methods.

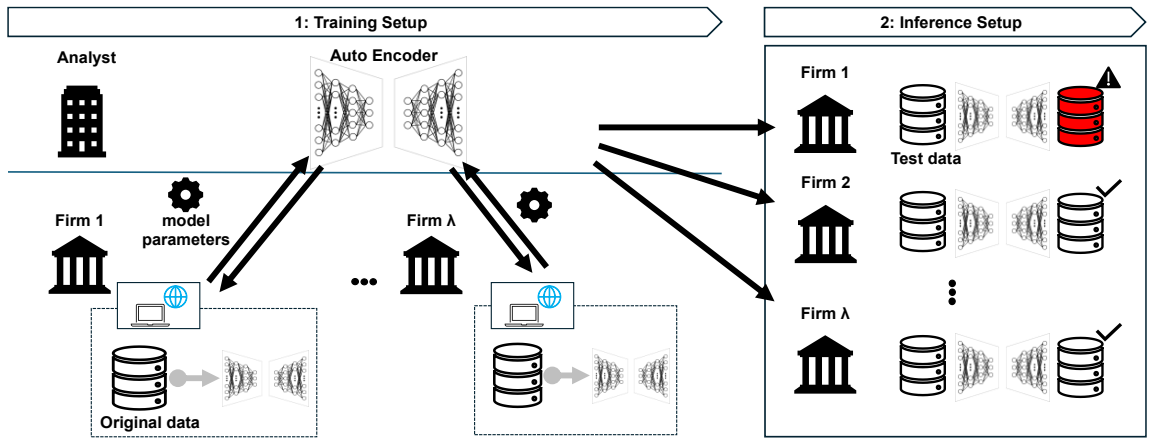

(a) model share-type FL based anomaly detection framework

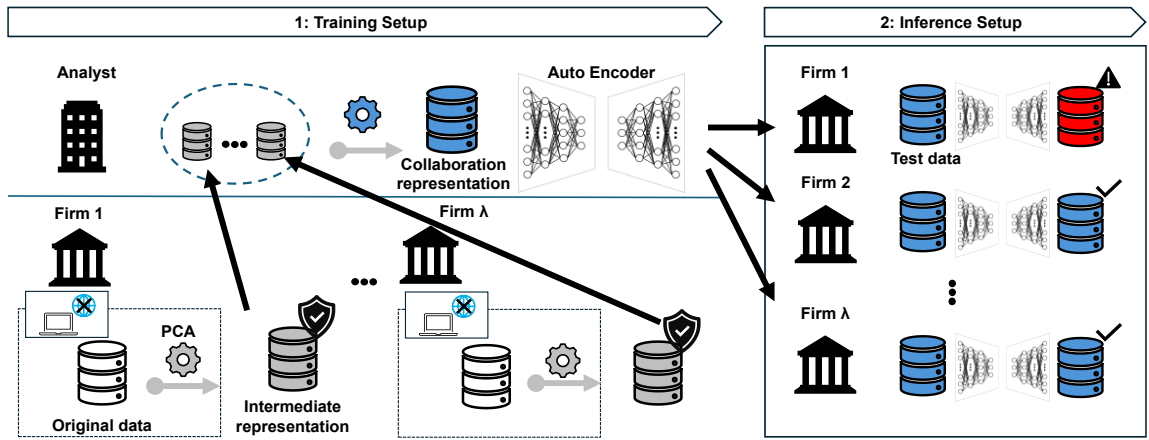

(b) DC analysis–based anomaly detection framework

Figure 1: Overview of the proposed method and comparison with existing model-sharing federated learning methods.

When integrating and analyzing journal entry data owned by different audit firms—or by separate divisions within the same firm—the types of accounts used and the scale of transaction amounts often differ substantially across entities. These distributional differences reflect the diversity of business operations and transaction characteristics among audited companies; such heterogeneity is commonly referred to as non-i.i.d. (non-independent and identically distributed) in the FL literature, where it can lead to performance degradation and unstable training. By contrast, i.i.d. (independent and identically distributed) describes the ideal scenario in which all entities share the same data distribution. In this study, the effectiveness and robustness of the proposed method are assessed under both i.i.d. and non-i.i.d. conditions to mimic real-world deployment environments.

The novelty of this study lies in the first application of DC analysis to unsupervised learning and the proposal of a journal entry anomaly detection framework that can build models with a single round of communication. Furthermore, additional innovation is demonstrated by employing journal entry data distributed across real organizations and comparative evaluations are conducting with existing FL methods in experiments that reflect actual distribution scenarios. By targeting the financial auditing domain, this research aims for practical application through the design and evaluation of an intelligent system capable of secure and high-precision anomaly detection in distributed environments. Our main contributions are as follows:

- We propose an anomaly detection framework based on data collaboration analysis, a non-model share-type FL approach. This framework enables collaborative model training in a single communication round—without connecting devices holding raw data to any external network—making it suitable for highly confidential journal entry data.
- Experiments using both synthetic and real data demonstrate that the proposed method outperforms models constructed by a single organization.
- We design experimental evaluations that reflect real-world non-i.i.d. conditions by using journal entry data distributed across multiple organizations. Our results show that the proposed method maintains high detection performance even under such heterogeneous settings. In particular, for local anomalies—which are relatively hard to detect and carry high fraud risk—our approach consistently outperforms existing model share-type FL methods.

The remainder of this paper is organized as follows. Section 2 reviews prior work on anomaly detection for journal entry data and on anomaly detection using FL. Section 3 describes the fundamentals of FL and DC analysis. Section 4 presents the proposed method. Section 5 reports the experimental results obtained on both synthetic and real-world datasets. Finally, Section 6 offers concluding remarks and outlines directions for future work.

## 2. Related Work
### 2.1. Anomaly Detection in Auditing

With the advent of Enterprise Resource Planning (ERP) systems and the resulting increase in data volume, anomaly detection in journal entry data has emerged as an important research topic in the domain of accounting and auditing research (Schreyer et al., 2017). Learning paradigms for anomaly detection are generally classified as unsupervised, supervised, or semi-supervised (Hilal et al., 2022). In the auditing domain, however, extensive labeling of journal entry data is practically infeasible, and labeled examples are scarce (Duan et al., 2025). Consequently, label-free approaches have become essential, and numerous unsupervised learning-based methods for anomaly detection in journal entries have been proposed.

Anomaly detection for journal entry data can be classified into three types on the basis of how the data are handled (Lu & Wu, 2025). The first type treats journal entry data as time series and applies time series anomaly detection methods. For example, Oliveira and Meira. (2006) regard transaction records in accounting systems as time series, employ a neural network forecasting model with robust confidence intervals to set thresholds for prediction errors, and flag future observations falling outside these intervals as anomalies. Similarly, Yoon et al. (2021) proposed an audit system with a three-tier architecture—comprising extraction of nonroutine transactions, detection of internal control violations, and identification of statistical anomalies—and, in the statistical anomaly layer, aggregate journal entries weekly and monthly, forecast them using a time series model, and evaluate deviations from expected values as anomalies.

The second type treats journal entry data as tabular data. No et al. (2019) proposed the Multidimensional Audit Data Selection (MADS) framework, which identifies outliers across multiple criteria and ranks transactions so that auditors can focus on the most significant cases. Wei et al. (2024) introduce a Multilevel Outlier Detection Framework that combines the local outlier factor (LOF) for local density evaluation, the Z-score for numeric outlier detection, and K-Modes clustering for categorical variables, demonstrating its ability to uncover diverse anomaly patterns in journal entries without labels. Schreyer et al. (2017) represented one of the earliest applications of DL to journal entry anomaly detection, using an autoencoder and reported that it achieved a higher AUC–ROC than other unsupervised methods. That study spurred further research on DL-based anomaly detection (Schreyer et al., 2019; Zupan et al., 2020; Müller et al., 2022). For example, Zupan et al. (2020) combined a Variational Autoencoder (VAE) with Long Short-Term Memory (LSTM), with the VAE targeting anomalies in account codes and the LSTM addressing anomalies in transaction amounts. Through experiments on multiyear journal entry data obtained from actual companies; they demonstrated the effectiveness of these two models in detecting anomalies in data from the most recent fiscal year.

Recently, graph-based approaches have also been introduced for anomaly detection in journal entry data. For example, Sotiropoulos et al. (2023) proposed ADAMM, a general anomaly detection model that integrates multiplex graph structures and metadata, and demonstrated its effectiveness on datasets including journal entries. Further, Huang et al. (2024) represented journal entry data as directed graphs and developed a transaction-level anomaly detection method via representation learning using a variational graph autoencoder (VGAE) that combines a graph neural network with a variational autoencoder.

## 2.2. Federated Learning and Anomaly Detection

FL frameworks can be classified into two primary categories—model share-type and non-model share-type approaches—e.g., DC analysis (Imakura et al., 2021b). In model share-type FL, each organization trains its own model on its client data, and the aggregated model is built by sharing and combining these model parameters. For example, the representative model share-type method FedAvg updates the aggregated model by computing the simple average of each organization's model weights. By contrast, DC analysis consolidates intermediate representations obtained via dimensionality reduction instead of sharing the models themselves and subsequently trains an aggregated model based on these representations.

FL has been widely adopted for anomaly detection tasks. For example, Zheng et al. (2020) target credit card transaction data and proposed a federated meta-learning framework that enables financial organizations to build few-shot learning–based fraud-detection models while keeping their data locally. Reddy et al. (2024) developed a model share-type FL framework using the Federated Averaging (FedAvg) algorithm, which combines a hybrid optimization algorithm with DL models so that banks can collaboratively train credit card fraud-detection models with high accuracy while preserving privacy. Tang and Liang (2024) converted transaction data into graph structures and integrated a GNN with FL—also based on FedAvg—leveraging locally constructed transaction graphs to achieve high-performance fraud detection in distributed environments. Moreover, Imakura et al. (2021b) applied DC analysis to anomaly detection using synthetic data and various open datasets. Their experiments demonstrated the effectiveness of DC analysis in anomaly detection and identified the need to apply DC analysis-based detection methods to more practically distributed data as a future goal.

To the best of our knowledge, only Schreyer et al. (2022) have applied FL to anomaly detection in journal entry data. They introduced a method that uses FedAvg to detect anomalies in journal entries and conducted experiments on municipal payment data with artificially inserted anomalies. Although this dataset was originally obtained from a single organization, it artificially simulates a multiorganization setup, thereby demonstrating that FL can integrate journal entry data obtained from multiple organizations and detect anomalies.

Despite the wealth of research on anomaly detection in journal entry data, most existing studies assume a single organization's dataset and thus do not address the distributed, multiorganization environment encountered in practice. Schreyer et al. (2022) proposed a method that integrates journal entry data from multiple companies via model share-type FL. However, their approach assumes that devices storing raw data remain connected to external networks and suffers from high communication costs. Moreover, to date, no study has evaluated anomaly detection under non-i.i.d. conditions using actual journal entry data collected from multiple organizations. To fill these gaps, this study proposes an anomaly detection method that treats journal entry data as tabular data and leverages DC analysis. We further validate its effectiveness and practicality by conducting experiments on real multiorganization journal entry datasets that simulate non-i.i.d. environments.

## 3. Preliminary
### 3.1. Federated Averaging

In this study, we adopt Federated Averaging (FedAvg), the canonical FL algorithm introduced by McMahan et al. (2017) and employed by Schreyer et al. (2022), as one of the baselines for comparison with our proposed method. FedAvg is a distributed learning technique in which each organization retains its raw data locally and participates in training by transmitting only model parameters. The basic workflow is as follows:

1. Initialization

The server initializes the aggregated model parameters as $w^{(0)}$, where $w^{(t)}$ denotes the aggregated model parameters in round $t$.

2. Local Training

For each communication round $t = 1, 2, \ldots, T$, the server sends the latest aggregated model parameter, $w^{(t)}$, to each client. Client $k$ adopts $w^{(t)}$ as the initial parameter and performs a few epochs of local training on its client data, $D_k$, resulting in updated parameters $w_k^{(t)}$.

3. Global Aggregation

Each client returns its locally updated parameters $w_k^{(t)}$ to the server. The server subsequently aggregates them to create the next round of parameters $w^{(t+1)}$. In this study, a weighted average determined by the proportion of each client's data size $n_k$ relative to the total data size $n$ across all clients is used for this purpose:

$$w^{(t+1)} \leftarrow \sum_{k=1}^{K} \frac{n_k}{n} w_k^{(t)} \qquad (1)$$

Here, $n_k$ denotes the number of data points held by client $k$, and $n = \sum n_k$ denotes the total number of data points across all clients.

By iteratively alternating local training (Step 2) and global aggregation (Step 3) in each communication round, FedAvg enables the construction of a high-performance aggregated model without requiring clients to transmit their raw data to the server.

### 3.2. FedProx

In financial auditing, the use of journal entry data from multiple organizations inherently leads to a non-i.i.d. setting. Therefore, we include FedProx (Li et al., 2020)—a representative FL method designed for non-i.i.d. environments—as another baseline for comparison. FedProx extends FedAvg by adding a proximal term to each client's local optimization to mitigate instability caused by distribution heterogeneity. Specifically, each client solves the following:

$$F_k(w) + \frac{\mu}{2} \|w - w_t\|^2 \qquad (2)$$

where $F_k(w)$ is the local objective for client $k$, $w_t$ is the aggregated model at round t, and $\mu$ is a hyperparameter that controls the strength of the proximal term. The proximal term $\frac{\mu}{2}\|w - w_t\|^2$ serves to penalize each client's local update if it deviates too far from the current aggregated model. By limiting the extent of these deviations, FedProx reduces the variability in update directions across clients, thereby promoting more stable convergence even under non-i.i.d. data.

### 3.3. Data Collaboration Analysis

Data collaboration analysis (DC analysis), proposed by Imakura and Sakurai (2020), is a non–model share-type distributed data analysis method. Similar to FedAvg and FedProx, DC analysis comprises clients that hold local data and an analyst (server) that constructs the aggregated model. In this method, privacy is preserved by converting distributed raw data into intermediate representations before aggregation rather than sharing the data directly. An intermediate representation is obtained by applying a dimensionality-reduction function—such as principal component analysis (PCA) (Pearson, 1901) or locality preserving projection (LPP) (He & Niyogi, 2003)—which each organization may choose independently. Because the dimensionality-reduction functions are never shared, no organization can infer another's raw data without access to its specific function. Once the intermediate representations are aggregated at the analyst, they are transformed back into a collaboration representation, enabling integrated analysis.

From here, we outline the fundamentals of DC analysis. Note that DC analysis can enable collaboration not only among organizations that are horizontally partitioned (i.e., samples distributed across organizations) but also among those that are vertically partitioned (i.e., features distributed across organizations). In this paper, however, we focus exclusively on sample-direction (horizontal) collaboration. Let $c$ denote the number of collaborating organizations, and $X_i \in \mathbb{R}^{n_i \times m}$ ($0 < i \leq c$) denote the raw data owned by the $i$-th organization. We also define $X^{anc} \in \mathbb{R}^{r \times m}$ as the anchor data, where $m$ denotes the dimension of the features and $r$ denotes the sample size. Anchor data are shared among all organizations and used to create the transformation function $g_i$, which converts intermediate representations into collaboration representations. The simplest form of anchor data can be a random matrix; however, it can also be generated from public data or basic statistics via methods such as random sampling, low-rank approximations, or synthetic minority oversampling techniques (Imakura et al., 2021a; Imakura et al., 2023).

The DC analysis algorithm proceeds as follows. Each organization creates its own intermediate representation function $f_i$. The intermediate representation of $\tilde{X}$ is expressed as

$$\tilde{X}_i = f_i(X) \in \mathbb{R}^{n_i \times \tilde{m}_i}, (0 < \tilde{m}_i < m) \quad (3)$$

where $0 < \tilde{m}_i < m$ denotes the dimensions of the intermediate representation. Using the same function $f_i$, each organization performs dimensionality reduction on the anchor data:

$$\tilde{X}_i^{anc} = f_i(X^{anc}) \in \mathbb{R}^{r \times \tilde{m}_i} \quad (4)$$

Subsequently, $\tilde{X}_i$ and $\tilde{X}_i^{anc}$ are shared with the analyst to create a collaboration representation.

Assuming that the mapping function $g_i$ from intermediate representations to collaboration representations is a linear transformation,

$$\hat{X}_i = g_i(\tilde{X}_i) = \tilde{X}_i G_i, \hat{X}_i^{anc} = g_i(\tilde{X}_i^{anc}) = \tilde{X}_i^{anc} G_i \tag{5}$$

These transformations can be determined by solving a least-squares problem using singular value decomposition (SVD) (Imakura & Sakurai, 2020). Let the low-rank approximation of the matrix obtained by horizontally concatenating $\tilde{X}_i^{anc}$ be

$$[\tilde{X}_1^{anc}, \tilde{X}_2^{anc}, \ldots, \tilde{X}_c^{anc}] = [U_1, U_2] \begin{bmatrix} \Sigma_1 & O \\ O & \Sigma_2 \end{bmatrix} \begin{bmatrix} V_1^T \\ V_2^T \end{bmatrix} \simeq U_1 \Sigma_1 V_1^T \tag{6}$$

Then, the transformation matrix, $G_i$, is obtained as

$$G_i = (\tilde{X}_i^{anc})^\dagger U_1 C \tag{7}$$

Here, † denotes the Moore–Penrose pseudoinverse, and $\|\cdot\|_F$ denotes the Frobenius norm. Moreover, $\Sigma_1 \in \mathbb{R}^{\widehat{m} \times \widehat{m}}$ is a diagonal matrix, $U_1$ and $V_1$ are orthogonal matrices, and $C \in \mathbb{R}^{\widehat{m} \times \widehat{m}}$ is an invertible matrix. Through these steps, the analyst obtains the collaboration representation

$$\hat{X} = [\hat{X}_1^T, \hat{X}_2^T, \ldots, \hat{X}_c^T]^T \in \mathbb{R}^{n \times \widehat{m}} \tag{8}$$

This collaboration representation can then be used for classification tasks, predictive modeling, or other forms of analysis.

## 4. Method

### 4.1. Autoencoder

In this study, an autoencoder is adopted as the anomaly detection model (Schreyer et al., 2017; Schultz & Tropmann, 2020; Schreyer et al., 2022). An autoencoder consists of two networks, an encoder and a decoder, that jointly learn to compress input data into a latent space and then reconstruct it back into the original space. Specifically, the encoder $f_\phi$ maps each observed data sample, $x \in R^d$, to a latent representation $\mathbf{z} \in \mathbb{R}^k$, whereas the decoder $g_\theta$ subsequently reconstructs it.

$$\mathbf{z} = f_\phi(\mathbf{x}), \hat{\mathbf{x}} = g_\theta(\mathbf{z}) \tag{9}$$

where $\phi$ and $\theta$ denote the parameters of the encoder and decoder, respectively. The autoencoder is trained such that $\hat{\mathbf{x}}$ approximates $\mathbf{x}$ as closely as possible.

$$\arg\min_{\phi,\theta} \|\mathbf{x} - g_\theta(f_\phi(\mathbf{x}))\| \tag{10}$$

Using an autoencoder for anomaly detection typically involves two steps. First, the autoencoder is trained solely on normal data. Then, when new data containing potential anomalies are subsequently input into the trained autoencoder, the reconstruction error between the output and input is calculated. Samples exhibiting large reconstruction errors are considered to be deviations from the learned representation of normal data and are thus flagged as potential anomalies.

The autoencoder's layer architecture is tailored to each of the two datasets described below. For experiments on synthetic data, we employ an autoencoder with the following architecture: [input layer, 6, 4, 2, 4, 6, output layer]. For experiments on real journal entry data, following Schreyer et al. (2022), we use an autoencoder with the following architecture: [input layer, 128, 64, 32, 16, 8, 4, 8, 16, 32, 64, 128, output layer].

## 4.2. Proposed Method

In this section, we describe our proposed anomaly detection method for journal entry data using DC analysis. Notably, our approach does not require devices holding raw data to connect to any external network, and it requires only a single communication round to integrate data. Our proposed method consists of four steps.

1. Creation of Intermediate Representations

First, each organization uses its historical, audited normal journal entry data to generate intermediate representations via a dimensionality reduction function $f$ and shares these representations with the analyst. Because the intermediate representations alone do not permit exact reconstruction of the raw data, confidentiality is preserved. Recovering the original data from the intermediate representations would require access to either the dimensionality reduction function $f$ itself or matched pairs of raw and projected features to approximate an inverse transformation. In our proposed method, only the projected features are shared externally, including with the analyst. This ensures that, even if the analyst were an adversary, inferring the original journal entries precisely from the shared representations would be infeasible.

In this study, we adopt PCA for dimensionality reduction. To apply PCA to journal entry data, we first preprocess the data via one-hot encoding of categorical variables and normalization of continuous variables. We then perform PCA on the preprocessed entries to obtain reduced-dimensional representations. The same PCA mapping is applied to the anchor data. Specifically, we set the target dimensionality $\widetilde{m}_i = m - 1$, and the anchor data consist of a random matrix with values uniformly sampled from 0 to 1.

2. Construction of Collaboration Representations

The analyst uses the intermediate representations of the anchor data collected from each organization to construct $G_i$ according to equation (7). Next, $G_i$ is employed to generate the collaboration representation $\hat{X}_i$ as defined in equation (5). Finally, these are integrated via equation (8) to train the autoencoder.

3. Autoencoder Training

The analyst trains the autoencoder (as described in Section 4.1) using the integrated representation $\hat{X}$. Rectified linear unit (ReLU) activations are applied to all the hidden layers, and identity activation is used for the output layer. The mean squared error (MSE) loss function is employed. By minimizing the MSE, the model parameters are updated to accurately reconstruct normal patterns—yielding low reconstruction errors—while producing higher reconstruction errors for unseen or anomalous patterns.

4. Anomaly Detection on Test Data

Finally, we perform anomaly detection on test data that may contain anomalous samples using the trained autoencoder. The analyst sends $G_i$ and the trained autoencoder to each organization. Given test data $Y_i$, each organization applies the same dimensionality reduction function used on the training data $X_i$ to produce $\tilde{Y}_i$. Subsequently, $\hat{Y}_i$ is generated using $G_i$ and input into the trained autoencoder for anomaly detection. Any journal entry with a high degree of reconstruction error is examined further by auditors, if necessary. The pseudocode for the proposed method is provided in the "Proposed Method" Algorithm.

| **Algorithm proposed method** | |
|---|---|
| | Input: $X_i \in \mathbb{R}^{n_i \times m}$ individually |
| | Output: reconstruction errors $\|\hat{Y}_i - h(\hat{Y}_i)\|$ |
| **I.** | **Creation of Intermediate Representations** |
| **1:** | Generate $X^{anc}$ and share to all organizations |
| **2:** | User $i$ Generate $f_i$ |
| **3:** | Construct $\tilde{X}_i = f_i(X_i), \tilde{X}_i^{anc} = f_i(X^{anc})$ |
| **4:** | Share $\tilde{X}_{k,l}, \tilde{X}_{k,l}^{anc}$ to analyzer |
| **II.** | **Construction of Collaboration Representations** |
| **5:** | Analyst obtains $\tilde{X}_{k,l}, \tilde{X}_{k,l}^{anc}, Y_k$ for all user $i$ |
| **6:** | Construct $g_k$ from $\tilde{X}_i^{anc}$ for all $i$ |
| **7:** | Construct $\hat{X}_i = g_i(\tilde{X}_i)$ for all $i$ and set $\hat{X}$ |
| **III.** | **Autoencoder Training** |
| **8:** | Construct $h$ by $\hat{X} = h(\hat{X})$ |
| **IV.** | **Anomaly Detection on Test Data** |
| **9:** | Share $h, G_i$ to user $i$ for all users |
| **10:** | User $i$ detect anomalies by $h(\hat{Y}_i) = h\left(g_i(f_i(Y_i))\right)$ |

## 5. Experiments

### 5.1. Experiment Settings

**Datasets.** We used two types of datasets: a synthetically generated simple dataset and a real-world journal entry dataset obtained from eight organizations. For both datasets, we generated two types of anomalies—global anomalies and local anomalies—following Schreyer et al. (2022). Note that, in this paper, the terms "global anomaly" and "local anomaly" are used based on the definitions in Breunig et al. (2000) and differ from the notions of "global model" and "local data" in the FL or DC analysis context. A global anomaly refers to a sample that contains extreme values when viewed against the entire dataset; it can be regarded as detecting outliers in individual attributes (Schreyer et al., 2017). These anomalies often correspond to unintentional errors, are relatively easy to detect, and carry a high probability of being mistakes (Schreyer et al., 2022). By contrast, a local anomaly refers to a sample whose combination of attribute values is abnormal compared with its local neighborhood or density; it corresponds to detecting anomalies at the level of combinations of multiple attributes (Schreyer et al., 2017). These anomalies may indicate intentional deviations, are relatively difficult to discover, and carry a high fraud risk (Schreyer et al., 2022).

The synthetic dataset comprises three variables, (a, b, c). Variables a and b are categorical and take values from the set {0,1,2}, whereas c is a continuous variable, with values ranging in [0,1] (Figure 2). In the synthetic dataset, global anomalies occur when c is significantly larger or smaller than that in normal data (i.e., below 0.1 or above 0.9), whereas local anomalies involve either abnormal (a, b) combinations or anomalous (a, b, c) combinations. The training set consists of 1,600 normal samples, and the test set comprises 200 samples containing both normal and anomalous cases. We injected anomalies into the test set at rates of 25% (25 global and 25 local anomalies), 10% (10 global and 10 local), and 5% (5 global and 5 local) to evaluate the effectiveness of the proposed method.

For the i.i.d. setting, we randomly partitioned the 1,600 training samples into eight organizations, assigning 200 samples to each. In the non-i.i.d. setting, following Laridi et al. (2024), we applied K-means clustering to the normal data and allocated each cluster to one of the eight organizations, thereby creating heterogeneous data distributions across organizations (Figure 3).

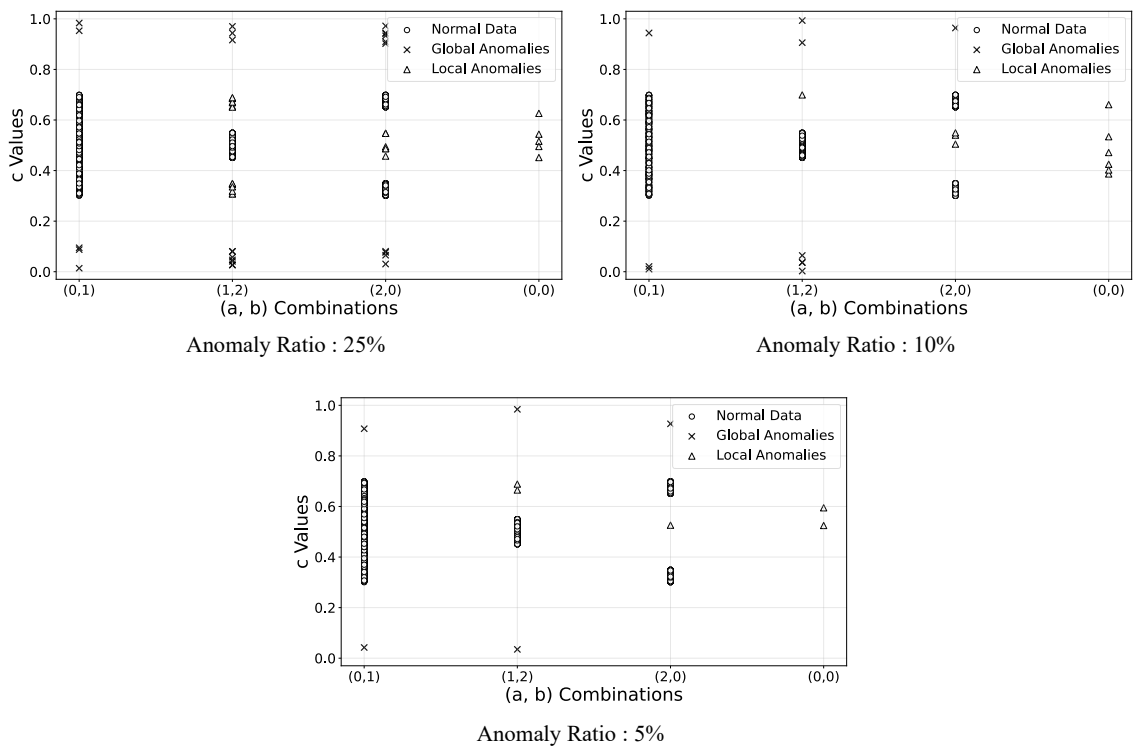

Figure 2: Distribution of synthetic data

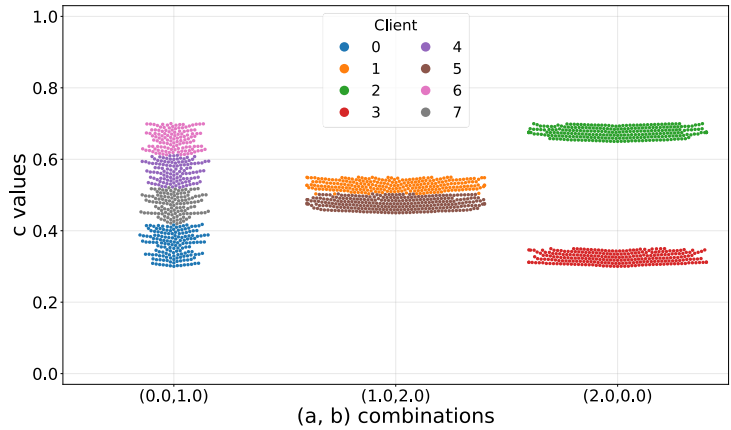

Figure 3: Distribution of synthetic data across clients after K-means-based non-i.i.d. setting.

The real journal entry dataset consists of multiple years of records from eight medical corporations in Japan. These data were provided by our collaborative research partners for research purposes and constitute confidential data from eight actual medical corporations. The dataset includes daily transaction records from 2016 to 2022, maintained according to double-entry bookkeeping rules. Basic descriptive statistics for the data are shown in Table 1. The features used in this study are the debit account, credit account, and transaction amount. Entries from 2016 through 2021 were used for model training, and the 2,737 entries from Clinic A in 2022 were used for testing. Since this dataset contains only normal entries, we generated synthetic anomalies and inserted them into the test set following Schreyer et al. (2022). The anomalous data comprise two types—global anomalies and local anomalies—mirroring the synthetic dataset. Global anomalies are defined as entries with transaction amounts that are extremely relative to the rest of the dataset; specifically, we multiply the amounts of the six largest entries in the normal data by factors of three to five. Local anomalies are generated in two ways: the first involves entries with anomalous account-pair combinations (for example, a debit account of Depreciation Expense and a credit account of Cash, a combination not seen in standard accounting practice), and the second involves altering the amounts of journal entries corresponding to regularly recurring transactions (such as rent or executive compensation). Details of the anomaly generation procedures are provided in the Appendix.

| Clinic | Number of accounts | | Transaction amounts (yen) | | | |
| --- | --- | --- | --- | --- | --- | --- |
| | Debit | Credit | Mean | Standard error | Median | Max |
| A | 69 | 53 | 399,135 | 1,488,541 | 49,043 | 17,632,000 |
| B | 74 | 61 | 263,080 | 1,577,547 | 8,898 | 23,496,050 |
| C | 69 | 57 | 630,734 | 2,742,157 | 73,016 | 91,670,000 |
| D | 68 | 56 | 377,151 | 1,342,280 | 24,893 | 12,868,437 |
| E | 70 | 64 | 391,185 | 2,346,763 | 10,150 | 100,000,000 |
| F | 72 | 63 | 465,457 | 1,440,960 | 23,180 | 13,335,000 |
| G | 74 | 67 | 743,130 | 3,003,221 | 52,827 | 50,000,770 |
| H | 68 | 51 | 596,171 | 1,910,126 | 34,560 | 26,946,396 |

Table 1: Numbers of accounts and statistics of transaction amounts.

In the i.i.d. setting for journal entry data, we first aggregate all entries obtained from the eight medical corporations and then randomly partition them into eight subsets, each treated as a separate organization. Thus, every organization holds data that are homogeneously and randomly sampled. In the non-i.i.d. setting, each medical corporation retains its original data distribution, conducting experiments in a manner closer to actual operations. In other words, we train while preserving differences in data volume and account type frequencies across organizations to evaluate performance under conditions reflective of real-world deployment. The number of samples held by each organization in both settings is shown in Table 2.

| Dataset | | 1 | 2 | 3 | 4 | 5 | 6 | 7 | 8 |
|---|---|---|---|---|---|---|---|---|---|
| Synthetic Data | i.i.d. | 200 | 200 | 200 | 200 | 200 | 200 | 200 | 200 |
| | non-i.i.d. | 272 | 244 | 162 | 261 | 119 | 289 | 125 | 128 |
| Journal entry data | i.i.d. | 12,780 | 12,780 | 12,780 | 12,780 | 12,780 | 12,780 | 12,780 | 12,779 |
| | non-i.i.d. | 17,070 | 22,978 | 10,708 | 11,230 | 14,984 | 8,525 | 8,133 | 8,611 |

Table 2: Number of samples per client for synthetic data and real journal entry data under i.i.d. and non-i.i.d. settings.

**Metrics.** In financial auditing, the goal of anomaly detection is twofold—to identify every anomalous journal entry (thereby maximizing recall) and to avoid excessive false alerts (thus minimizing false positives) (Schultz and Tropmann, 2020). To balance these competing requirements, the average precision (AP), derived from the precision-recall (PR) curve, is well suited as an evaluation metric. In this study, following Schreyer et al. (2022), we treat the reconstruction error from an autoencoder as an anomaly score, generate a PR curve by varying the error threshold, and calculate the area under this curve. Consequently, utilizing this metric, we comprehensively assess the ability of our approach to enhance recall while minimizing false positives.

**Baselines.** In this study, we compared the proposed method against the following four models:

- Individual Analysis (IA)

  A method that builds an anomaly detection model using only the data from a single organization. When training samples are insufficient, the model may be undertrained.

- Centralized Analysis (CA)

  A method that aggregates each organization's raw data and trains a model on the combined dataset. Although this approach theoretically achieves the highest performance, it requires direct sharing of confidential data and is therefore impractical for real-world deployment.

- FedAvg

  The representative model share-type FL method used by Schreyer et al. (2022).

- FedProx

  An extension of FedAvg designed for non-i.i.d. distributed data.

We evaluate the effectiveness of the proposed DC-based method by demonstrating that it outperforms IA and by comparing it with existing methods (FedAvg and FedProx) as well as the ideal but impractical CA.

The following experimental parameters were used: a batch size of 32 and a learning rate of 0.001. For IA, CA, and DC, we set the number of epochs to 200. For FedAvg and FedProx, following Bogdanova et al. (2020), we used 20 epochs per client and 10 aggregation rounds so that the total training effort matched that of IA/CA/DC. For IA, CA, FedAvg, and FedProx, the input to the autoencoder consisted of journal entry data that had been one‐hot encoded and normalized. Accordingly, the output‐layer activation functions and loss functions were chosen based on variable type: categorical variables use softmax activation and binary cross‐entropy loss, whereas continuous variables use linear activation and MSE loss. The total autoencoder loss is the sum of these component losses. All the experiments were implemented in Python via Keras and conducted on a machine equipped with a 13th Gen Intel® Core™ i7-13700KF CPU, an NVIDIA GeForce RTX 4060 laptop GPU, and 16 GB of RAM.

### 5.2. Results and Discussion

In experiments using both synthetic and real journal entry data, each setup was repeated 10 times with random initialization of the autoencoder parameters, and the mean performance was evaluated. $AP_{all}$ measures the overall detection performance across both anomaly types, whereas $AP_{global}$ and $AP_{local}$ assess the detection performance for global anomalies and local anomalies, respectively. The parameter $\lambda$ denotes the number of participating organizations: for $\lambda = 4$, data from four of the eight organizations are used, and for $\lambda = 8$, data from all eight organizations are used.

**Synthetic data**

The experimental results on the synthetic data are shown in Figures 4 and 5. First, compared with IA, our proposed method outperformed IA on all the metrics ($AP_{all}$, $AP_{global}$, and $AP_{local}$) under both i.i.d. and non-i.i.d. settings. This finding indicates that safely integrating data distributed across multiple organizations can yield a more effective anomaly detection model than training on a single organization's data alone. Under the i.i.d. setting, our method also outperformed FedAvg and FedProx in many scenarios. In particular, for $AP_{all}$, although FedAvg and FedProx exhibited marked performance degradation when the anomaly rate decreased from 25% to 5%, the decrease in performance for DC analysis was relatively modest. This suggests that our method remains effective even under the highly imbalanced conditions that are typical in auditing practices, where the proportion of anomalies is small. Looking at each anomaly type separately, for global anomalies, the proposed method achieved AP very close to that of CA, which assumes the ideal scenario of fully centralized data. For local anomalies, it outperformed FedAvg and FedProx in most cases and maintained performance close to that of CA.

In the non-i.i.d. setting, $AP_{all}$ tended to decrease overall compared with the i.i.d. setting, with FedAvg exhibiting a particularly large drop in performance. Although FedProx achieved higher $AP_{all}$ than FedAvg did, our proposed DC analysis method outperformed FedProx under many configurations. For global anomaly detection, DC analysis continued to maintain AP comparable to that of CA and demonstrated robust performance even under non-i.i.d. conditions. On the other hand, local anomaly detection proved more challenging overall, and AP decreased for all methods; however, DC analysis consistently outperformed FedAvg and FedProx and retained the performance closest to that of CA. These results indicate that the proposed method delivers superior anomaly detection performance compared with existing approaches across numerous conditions in both i.i.d. and non-i.i.d. settings and is particularly effective in the non-i.i.d. environments typical of real-world deployment. The detailed mean values and standard deviations for each condition are provided in the Appendix.

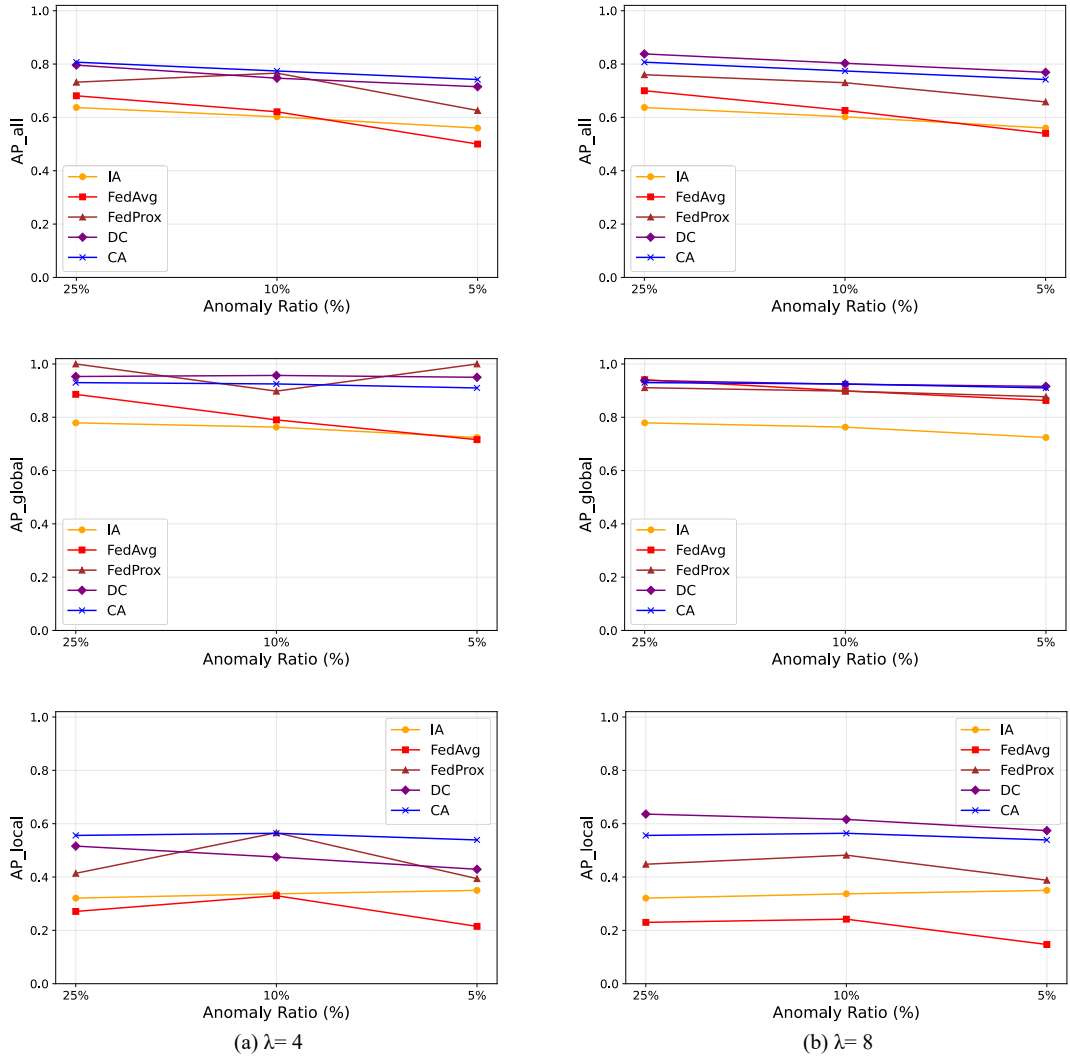

Figure 4: AP comparison of all the models on synthetic data under the i.i.d. setting.

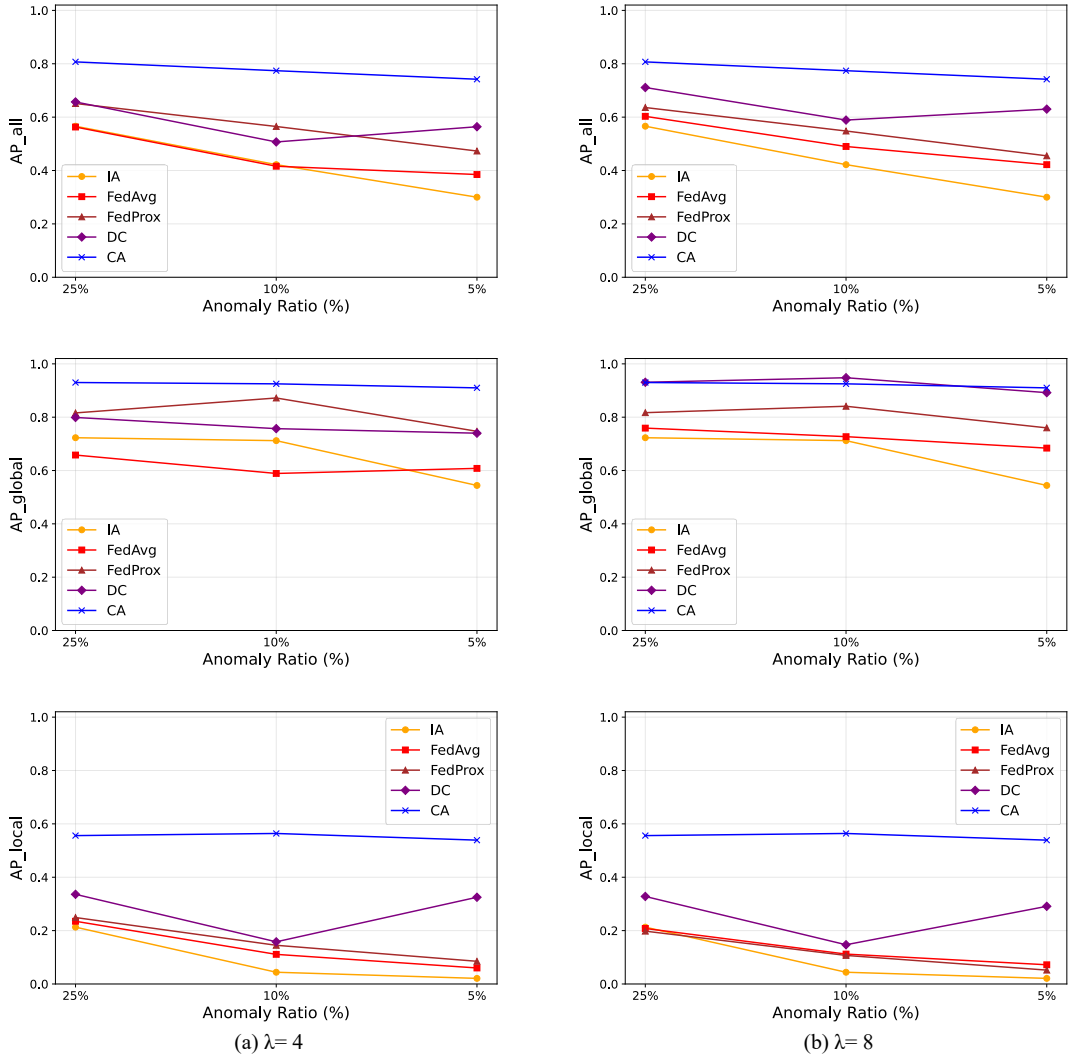

Figure 5: AP comparison of all the models on synthetic data under the non-i.i.d. setting.

**Real journal entry data**

The experimental results using the real journal entry data are shown in Table 3 and Table 4. Under every condition, the proposed method outperformed IA on all three AP metrics. This demonstrates that—even when real journal entry data are used—our method, which jointly trains on data from multiple organizations, yields a more effective model than one built on data from a single organization. In the i.i.d. setting, among the four methods other than CA, which is impractical because of confidentiality concerns, DC analysis achieved the highest or second-highest AP scores. Specifically, for $AP_{all}$ and $AP_{local}$, FedAvg attained the highest AP, with the proposed method ranking second. For $AP_{global}$, the proposed method achieved detection performance comparable to FedProx and surpassed FedAvg.

In the non-i.i.d. setting, although AP metrics generally declined, DC analysis achieved the highest detection performance under most conditions among the four methods, excluding centralized analysis (CA). In particular, DC analysis consistently outperformed FedAvg and FedProx on $AP_{all}$ and $AP_{local}$. For $AP_{global}$ delivered results very close to those of FedProx, which achieved the best performance. These findings are based on evaluations using journal entry data that reflect real auditing environments, indicating that the proposed method may be effective in practical deployments. Importantly, the non-i.i.d. setting captures differences in account distributions across organizations and business units, and the robustness of our method to such data heterogeneity is a key outcome.

| $\lambda$ | 4 | | | 8 | | |
|---|---|---|---|---|---|---|
| AP | $AP_{all}\uparrow$ | $AP_{global}\uparrow$ | $AP_{local}\uparrow$ | $AP_{all}\uparrow$ | $AP_{global}\uparrow$ | $AP_{local}\uparrow$ |
| IA | 0.496 (0.064) | 0.913 (0.097) | 0.196 (0.070) | 0.496 (0.064) | 0.913 (0.097) | 0.196 (0.070) |
| FedAvg | **0.585** (0.087) | 0.857 (0.138) | **0.326** (0.083) | **0.614** (0.035) | 0.813 (0.164) | **0.375** (0.074) |
| FedProx | 0.471 (0.056) | **0.990** (0.029) | 0.149 (0.065) | 0.491 (0.107) | <u>0.985</u> (0.046) | 0.191 (0.152) |
| DC | <u>0.513</u> (0.077) | **0.990** (0.021) | <u>0.198</u> (0.109) | <u>0.512</u> (0.037) | **1.000** (0.000) | <u>0.206</u> (0.063) |
| CA | 0.686 (0.060) | 0.958 (0.075) | 0.440 (0.099) | 0.686 (0.060) | 0.958 (0.075) | 0.440 (0.099) |

Table 3: AP comparison of all models on journal entry data under the i.i.d. setting. Values without parentheses represent the means, whereas those within parentheses indicate the standard deviation. Excluding CA, the best scores are shown in bold, and the second-best scores are underlined.

| $\lambda$ | 4 | | | 8 | | |
|---|---|---|---|---|---|---|
| AP | $AP_{all}\uparrow$ | $AP_{global}\uparrow$ | $AP_{local}\uparrow$ | $AP_{all}\uparrow$ | $AP_{global}\uparrow$ | $AP_{local}\uparrow$ |
| IA | 0.369 (0.008) | 0.824 (0.248) | 0.092 (0.019) | 0.369 (0.008) | 0.824 (0.248) | 0.092 (0.019) |
| FedAvg | 0.402 (0.118) | 0.689 (0.309) | <u>0.169</u> (0.058) | <u>0.491</u> (0.035) | 0.936 (0.078) | <u>0.187</u> (0.058) |
| FedProx | <u>0.424</u> (0.047) | **1.00** (0.000) | 0.087 (0.037) | 0.368 (0.013) | **1.00** (0.000) | 0.045 (0.008) |
| DC | **0.562** (0.034) | <u>0.998</u> (0.007) | **0.256** (0.056) | **0.495** (0.035) | <u>0.993</u> (0.021) | **0.188** (0.054) |
| CA | 0.686 (0.060) | 0.958 (0.075) | 0.440 (0.099) | 0.686 (0.060) | 0.958 (0.075) | 0.440 (0.099) |

Table 4: AP comparison of all models on journal entry data under the non-i.i.d. setting. The presentation style is consistent with that of Table 3.

From these results, the proposed method consistently outperforms IA and, in many cases, demonstrates superior anomaly detection performance compared with the existing methods FedAvg and FedProx. In particular, its efficacy under the non-i.i.d. setting—which closely resembles real-world operations—is remarkable; especially for local anomalies, which are relatively difficult to detect and carry high fraud risk, our method consistently achieves higher detection performance than the baselines do. These findings suggest that the proposed approach is effective in actual auditing scenarios where multiple organizations maintain distinct data distributions. However, the detection performance for local anomalies still shows a substantial gap compared with CA. We speculate that this degradation in performance could stem from the intermediate representations generated in our approach; as the data obtained via one-hot encoding of journal entries are highly sparse, the subsequent dimensionality reduction may not preserve sufficient information for generating an effective collaboration representation, ultimately reducing detection accuracy.

## 6. Conclusion

In this paper, we have proposed a framework that integrates journal entry data distributed across multiple organizations to construct an anomaly detection model with only a single communication round without requiring devices holding raw data to connect to the internet. Our method is based on DC analysis and builds an aggregated model by integrating only intermediate representations derived from each organization's data. The novelties of this study include (1) the first application of DC analysis to an unsupervised anomaly detection task, completing model construction in a single communication round, and (2) the use of real journal entry data provided by multiple organizations to conduct comparative evaluations against existing FL methods under non-i.i.d. settings. Experiments on both synthetic and real data demonstrated that the proposed method consistently outperforms models trained on a single organization's data and, under most conditions, exceeds the performance of FedAvg and FedProx. Our framework addresses the critical confidentiality challenges of AI deployment in auditing and holds promise for advancing practical AI applications in this domain.

Several challenges remain. First, the method of creating intermediate representations requires further refinement. In this study, we use PCA for dimensionality reduction; however, applying PCA to sparse data, such as journal entries, may result in insufficient preservation of information in collaboration representations, potentially degrading the performance of the anomaly detection model. Additionally, the effectiveness of preventing raw data inference through PCA-based dimensionality reduction on sparse data warrants further investigation. Second, validation using real-world anomalies remains. Although we followed Schreyer et al. (2022) in generating synthetic anomaly data, future work should evaluate the detection of actual fraudulent or misposted journal entries. Finally, journal entry records include additional information such as transaction dates, data-entry personnel, and journal descriptions that were not used in this study. The incorporation of these supplementary attributes is expected to enable the development of more practical and comprehensive anomaly detection methods for journal entry data.

## CRediT authorship contribution statement

**Sota Mashiko:** Conceptualization, Methodology, Software, Formal analysis, Investigation, Data curation, Writing – original draft, Visualization. **Yuji Kawamata:** Conceptualization, Methodology, Writing – review & editing, Visualization. **Tomoru Nakayama:** Methodology, Software, Writing – review & editing. **Tetsuya Sakurai:** Conceptualization, Writing – review & editing, Supervision, Project administration, Funding acquisition. **Yukihiko Okada:** Conceptualization, Methodology, Writing – review & editing, Supervision, Project administration, Funding acquisition.

## Acknowledgments

This study was supported by the Japan Society for the Promotion of Science Grants-in-Aid for Scientific Research (no. JP23K22166). We express our gratitude to the clinics for providing their confidential double-entry bookkeeping data for the experiments. We would like to thank Editage (www.editage.com) for English language editing.

## Declaration of generative AI in scientific writing

During the preparation of this work the authors used ChatGPT (OpenAI) to assist in translating parts of the manuscript from Japanese to English and to improve its readability. After using this tool, the authors reviewed and edited the content as needed and take full responsibility for the content of the published article.

# Appendix

## A. Details of Injected Anomalies in Journal Entry Data

Table A.1 presents the specific anomalous entries that were artificially generated for the experiments.

| Debit (Dr.) | Credit (Cr.) | Amount (yen) | Anomaly |
|---|---|---|---|
| Insurance accounts receivable | Sales (Insurance claim income) | 88,160,000 | global |
| Sales (Insurance claim income) | Insurance accounts receivable | 88,160,000 | global |
| Insurance accounts receivable | Sales (Insurance claim income) | 67,000,000 | global |
| Insurance accounts receivable | Sales (Insurance claim income) | 65,188,000 | global |
| Sales (Insurance claim income) | Insurance accounts receivable | 47,205,000 | global |
| Insurance accounts receivable | Sales (Insurance claim income) | 47,205,000 | global |
| Depreciation | Cash | 300,000 | local |
| Entertainment expenses | Salaries and allowances | 1,500 | local |
| Utilities expenses | Insurance accounts receivable | 1,900 | local |
| Accounts payable - trade | Advertising expenses | 30,000 | local |
| Communication expenses | Insurance accounts receivable | 2,100 | local |
| Miscellaneous expenses | Interest income | 2,000 | local |
| Sales (paid at one's own expense) | Taxes and dues | 400,000 | local |
| Accounts payable - trade | Insurance expenses | 20,000 | local |
| Income taxes - current | Short-term loans receivable | 49,200 | local |
| Deposits received | Construction in progress | 1,000 | local |
| Rents | Ordinary deposits | 6,270,000 | local |
| Membership fee | Ordinary deposits | 800,000 | local |
| Directors' compensations | Sundries | 62,650 | local |
| Depreciation | Accumulated depreciation | 2,920,000 | local |

Table A.1: List of anomalous journal entries

## B. Detailed Results of Experiments Using Synthetic Data

Tables B.1 to B.4 show the average precision (AP) scores and standard deviations obtained from the experiments using synthetic data. Each table corresponds to a different setting, varying by the number of participating clients ($\lambda$ = 4 or 8) and data distribution (i.i.d. or non-i.i.d.). For each setting, the tables report the means and standard deviations of $AP_{all}$, $AP_{global}$, and $AP_{local}$ across different anomaly ratios.

| ratio | 25% | | | 10% | | | 5% | | |
|---|---|---|---|---|---|---|---|---|---|
| AP | $AP_{all}$ | $AP_{global}$ | $AP_{local}$ | $AP_{all}$ | $AP_{global}$ | $AP_{local}$ | $AP_{all}$ | $AP_{global}$ | $AP_{local}$ |
| IA | 0.637 (0.102) | 0.779 (0.121) | 0.321 (0.102) | 0.602 (0.177) | 0.763 (0.173) | 0.337 (0.267) | 0.560 (0.161) | 0.724 (0.121) | 0.350 (0.291) |
| FedAvg | 0.681 (0.095) | 0.886 (0.158) | 0.271 (0.135) | 0.621 (0.189) | 0.790 (0.281) | 0.330 (0.268) | 0.500 (0.189) | 0.716 (0.305) | 0.215 (0.219) |
| FedProx | 0.781 (0.085) | **1.000** (0.000) | 0.414 (0.230) | **0.766** (0.230) | 0.898 (0.118) | **0.566** (0.194) | 0.717 (0.133) | **1.000** (0.000) | 0.394 (0.276) |
| DC | **0.796** (0.126) | 0.953 (0.072) | **0.516** (0.330) | 0.747 (0.166) | **0.957** (0.058) | 0.475 (0.343) | 0.715 (0.148) | 0.950 (0.116) | **0.429** (0.337) |
| CA | 0.807 (0.128) | 0.930 (0.135) | 0.556 (0.321) | 0.774 (0.157) | 0.925 (0.166) | 0.564 (0.355) | 0.742 (0.179) | 0.910 (0.161) | 0.539 (0.379) |

Table B.1: AP comparison of all models on synthetic data under the i.i.d. setting with 4 participating clients. Values without parentheses represent the means, whereas those within parentheses indicate the standard deviation. Excluding CA, the best scores are shown in bold, and the second-best scores are underlined.

| ratio | 25% | | | 10% | | | 5% | | |
|---|---|---|---|---|---|---|---|---|---|
| AP | $AP_{all}$ | $AP_{global}$ | $AP_{local}$ | $AP_{all}$ | $AP_{global}$ | $AP_{local}$ | $AP_{all}$ | $AP_{global}$ | $AP_{local}$ |
| IA | 0.637 (0.102) | 0.779 (0.121) | 0.321 (0.102) | 0.602 (0.177) | 0.763 (0.173) | 0.337 (0.267) | 0.560 (0.161) | 0.724 (0.121) | 0.350 (0.291) |
| FedAvg | 0.700 (0.040) | <u>0.941</u> (0.109) | 0.230 (0.092) | 0.626 (0.082) | <u>0.899</u> (0.175) | 0.242 (0.249) | 0.540 (0.008) | 0.863 (0.229) | 0.147 (0.163) |
| FedProx | <u>0.760</u> (0.112) | 0.911 (0.128) | <u>0.448</u> (0.229) | <u>0.730</u> (0.129) | 0.898 (0.217) | <u>0.482</u> (0.281) | <u>0.658</u> (0.156) | <u>0.877</u> (0.222) | <u>0.388</u> (0.273) |
| DC | **0.838** (0.149) | **0.938** (0.087) | **0.636** (0.344) | **0.803** (0.182) | **0.924** (0.095) | **0.616** (0.359) | **0.769** (0.209) | **0.916** (0.133) | **0.574** (0.382) |
| CA | 0.807 (0.128) | 0.930 (0.135) | 0.556 (0.321) | 0.774 (0.157) | 0.925 (0.166) | 0.564 (0.355) | 0.742 (0.179) | 0.910 (0.161) | 0.539 (0.379) |

Table B.2: AP comparison of all models on synthetic data under the i.i.d. setting with 8 participating clients. The presentation style is consistent with that of Table B.2.

| ratio | 25% | | | 10% | | | 5% | | |
|---|---|---|---|---|---|---|---|---|---|
| AP | $AP_{all}$ | $AP_{global}$ | $AP_{local}$ | $AP_{all}$ | $AP_{global}$ | $AP_{local}$ | $AP_{all}$ | $AP_{global}$ | $AP_{local}$ |
| IA | 0.566 (0.032) | 0.723 (0.048) | 0.213 (0.045) | 0.422 (0.026) | 0.712 (0.053) | 0.044 (0.002) | 0.300 (0.043) | 0.544 (0.086) | 0.021 (0.001) |
| FedAvg | 0.563 (0.074) | 0.658 (0.115) | 0.235 (0.053) | 0.416 (0.091) | 0.589 (0.139) | 0.111 (0.057) | 0.385 (0.051) | 0.608 (0.075) | 0.060 (0.014) |
| FedProx | <u>0.651</u> (0.046) | **0.816** (0.077) | <u>0.249</u> (0.050) | **0.565** (0.036) | **0.872** (0.071) | <u>0.145</u> (0.030) | <u>0.473</u> (0.083) | **0.747** (0.127) | <u>0.085</u> (0.055) |
| DC | **0.657** (0.052) | <u>0.799</u> (0.115) | **0.336** (0.037) | <u>0.507</u> (0.088) | <u>0.757</u> (0.183) | **0.158** (0.016) | **0.564** (0.092) | <u>0.740</u> (0.144) | **0.325** (0.077) |
| CA | 0.807 (0.128) | 0.930 (0.135) | 0.556 (0.321) | 0.774 (0.157) | 0.925 (0.166) | 0.564 (0.355) | 0.742 (0.179) | 0.910 (0.161) | 0.539 (0.379) |

Table B.3: AP comparison of all models on synthetic data under the non-i.i.d. setting with 4 participating clients. The presentation style is consistent with that of Table B.2.

| ratio | 25% | | | 10% | | | 5% | | |
|---|---|---|---|---|---|---|---|---|---|
| AP | $AP_{all}$ | $AP_{global}$ | $AP_{local}$ | $AP_{all}$ | $AP_{global}$ | $AP_{local}$ | $AP_{all}$ | $AP_{global}$ | $AP_{local}$ |
| IA | 0.566 (0.032) | 0.723 (0.048) | 0.213 (0.045) | 0.422 (0.026) | 0.712 (0.053) | 0.044 (0.002) | 0.300 (0.043) | 0.544 (0.086) | 0.021 (0.001) |
| FedAvg | 0.603 (0.062) | 0.759 (0.087) | <u>0.208</u> (0.065) | 0.490 (0.077) | 0.727 (0.145) | <u>0.112</u> (0.003) | 0.422 (0.072) | 0.684 (0.113) | <u>0.072</u> (0.062) |
| FedProx | <u>0.636</u> (0.046) | <u>0.817</u> (0.090) | 0.198 (0.028) | <u>0.548</u> (0.058) | <u>0.841</u> (0.122) | 0.107 (0.022) | <u>0.455</u> (0.064) | <u>0.760</u> (0.131) | 0.052 (0.005) |
| DC | **0.711** (0.026) | **0.931** (0.066) | **0.328** (0.064) | **0.589** (0.037) | **0.948** (0.083) | **0.147** (0.008) | **0.630** (0.079) | **0.892** (0.102) | **0.291** (0.122) |
| CA | 0.807 (0.128) | 0.930 (0.135) | 0.556 (0.321) | 0.774 (0.157) | 0.925 (0.166) | 0.564 (0.355) | 0.742 (0.179) | 0.910 (0.161) | 0.539 (0.379) |

Table B.4: Table B.3: AP comparison of all models on synthetic data under the non-i.i.d. setting with 8 participating clients. The presentation style is consistent with that of Table B.2.